\documentclass[letterpaper, 10 pt, conference]{ieeeconf}  

\IEEEoverridecommandlockouts                              

\usepackage{cite}
\usepackage{graphics} 
\usepackage{epsfig} 
\usepackage{mathptmx} 
\usepackage{times} 
\usepackage{amsmath} 
\usepackage{amssymb}  
\usepackage{color}
\usepackage{blkarray}
\usepackage{mathtools}
\usepackage{pbox}
\usepackage{breqn}
\usepackage{algorithmicx}
\usepackage{algorithm}
\usepackage[noend]{algpseudocode}
\usepackage{stfloats}
\usepackage{subfig}
\usepackage{svg}
\usepackage{url}
\usepackage{tikz}
\usepackage{array}
\usepackage{varwidth}
\usepackage{color}
\usepackage{cite}
\usepackage{amsmath}
\usepackage{graphicx}
\usepackage{amsfonts}
\usepackage{balance}
\usepackage{array}
\usepackage{makecell}
\usetikzlibrary{arrows, arrows.meta, shapes,
automata, backgrounds, petri, matrix, calc, shapes}

\tikzset{
  treenode/.style = {shape=rectangle, rounded corners,
                     draw, anchor=center,
                     text width=6.5em, align=center,
                     top color=white, bottom color=gray!30,
                     inner sep=1ex},
  decision/.style = {treenode, diamond, inner sep=0pt},
  root/.style     = {treenode
  },
  env/.style      = {treenode},
  finish/.style   = {root, bottom color=green!40},
  dummy/.style    = {}
}

\makeatletter
\def\BState{\State\hskip-\ALG@thistlm}
\makeatother

\DeclareMathOperator*{\argmax}{arg\,max}

\title{\LARGE \bf
A novel Skill-based Programming Paradigm based on Autonomous Playing and Skill-centric Testing
}

\author{Simon Hangl, Andreas Mennel and Justus Piater$^{*}$
\thanks{$^{*}$All authors are with the Department of Informatics, University of Innsbruck, Austria
        {\tt\small first.last@uibk.ac.at}}%
}

\begin{document}

\maketitle
\thispagestyle{empty}
\pagestyle{empty}
\begin{abstract}
We introduce a novel paradigm for robot programming with which we aim
to make robot programming more accessible for unexperienced users.
In order to do so we incorporate two major components in one single framework:
\emph{autonomous skill acquisition by robotic playing} and \emph{visual programming}.
Simple robot program skeletons solving a task for
one specific situation, so-called \emph{basic behaviours}, are provided by the user.
The robot then learns how to solve the same task in many different
situations by \emph{autonomous playing} which reduces the barrier for
unexperienced robot programmers.
Programmers can use a mix of \emph{visual programming} and \emph{kinesthetic teaching}
in order to provide these simple program skeletons.
The robot program can be implemented interactively by programming parts with
visual programming and kinesthetic teaching.
We further integrate work on experience-based skill-centric robot software
testing which enables the user to continuously test implemented skills without
having to deal with the details of specific components.
\end{abstract}
\section{Introduction}
Despite large progress in the community to make robot programming more accessible
to a larger public, it is still very hard for beginners to get started.
Software projects in robotics consist of a complex interplay of many components
such as robot control, object detection, machine learning, path planning and many more.
Additionally, skill acquisition in robotics is a very hard problem which is still
not solved yet.
Due to the diversity of the state spaces and the high
number of the controlled degrees of freedom it is very hard to design
well-generalising controllers.
This problem is often solved by introducing large task-specific priors
designed by engineers who are experts in a certain subfield.

This is clearly not an option if the goal is to enable beginners
to design simple robotic skills.
Another possibility is to design abstract visual programming languages
with which the users can easily program a skill for a limited range of situations.
Provided with such a simple program, the robot can add the rest, e.g. the generality
by autonomous playing.

In previous work work we introduced an approach for (semi-) autonomous
skill acquisition in which programmers provide so-called \emph{basic behaviours},
that solve a task in a narrow range of situations \cite{hangl2017tro, Hangl-2016-IROS}.
This range is extended by autonomous playing,
where sequences of behaviours are used to prepare the environment
such that the basic behaviour can be applied successfully.
We further introduced a novel approach for autonomous skill-centric
testing which enables even non-experts to automatically run tests
on robot software without the need for designing unit tests \cite{Hangl-2017-IROS}.

We propose a novel programming paradigm based on our previous
work on autonomous robots \cite{Hangl-2016-IROS, hangl2017tro, Hangl-2017-IROS}.
We extend our system to support visual programming.
Developers can implement basic behaviours via drag and drop for
a narrow range of situations.
Further, developers can interactively add parts of the basic behaviour
by kinesthetic teaching.
After providing the basic behaviour the rest, i.e. how to generalise to more
situations, can be learned autonomously by the robot.

Further, we embedded autonomous testing to our programming framework
in which buggy functions can be identified without deep knowledge of
robotic systems.
This can help to pinpoint problems which can be fixed either by the developers
themselves if they have the required knowledge, or by experts that can
be consulted based on the output.
\section{Contribution}
Our contribution is an integrated platform for robot development
based on technology for autonomous robotics.
We combine previous work on autonomous skill acquisition and robot software
testing \cite{Hangl-2016-IROS, hangl2017tro, Hangl-2017-IROS} in a unified framework.
We further extend our framework by a visual programming
capability in which a basic behaviours can be implemented in a simple way.
Skills implemented this way can then be tested autonomously without requiring
the user to be an expert.

We provide an abstract architecture and an open source
implementation\footnote{\url{https://github.com/shangl/kukadu}}.
We demonstrate that our approach can be used to program simple robotic skills
by visual programming and to make them generalise by autonomous robotic playing.
This simplifies the implementation process and makes robot programming accessible
to non-experts and even end-users.
\section{Related work}
Our framework aims to contribute to the field of end-user development
\cite{lieberman2006end}.
Domain-specific end-user programming is a powerful tool in order to
enable users to develop applications for their own purpose.
Eco-systems for end-user programming have to fulfill specific
properties to ensure that end-users are able to understand them
properly \cite{ko2004six}.
Frameworks and programming languages for end-users need to be simple
but powerful enough to implement useful programs.
One way is to follow a visual programming paradigm using abstract symbols
in which the symbols are tailored to the specific domain.
Popular examples are the \emph{scratch} framework \cite{resnick2009scratch},
\emph{Snap!}\footnote{\url{http://snap.berkeley.edu/}} or
\emph{Blockly}\footnote{\url{https://developers.google.com/blockly/}}.

Visual programming received some attention in robotics in recent years
\cite{kim2007programming, jackson2007microsoft, nguyen2013ros} and it was
shown that properly designed visual programming interfaces can make
robot programming even accessible to children \cite{riedo2013thymio}.
Each of these methods incorporates certain assumptions in order to ensure
simplicity and the required generality.
Jackson published on the Microsoft robotics studio (MSRS) \cite{jackson2007microsoft}
which aims to be a general purpose robotics platform.
Similar to the software presented in this paper, interfaces
for hardware are defined in order to provide an abstraction from the raw hardware.
This hardware can be used in a webservice-like structure, where distributed
webservices implement certain skills and behaviours.
Similar to our approach, skills and behaviours can be coordinated by visual programming.
Kim and Jeon designed a visual programming language based on
LabVIEW\footnote{\url{http://www.ni.com/de-at/shop/labview.html}} and the MSRS
for Lego mindstorm robots \cite{kim2007programming}.

Nguyen et al. \cite{nguyen2013ros} follow a different approach with their system
called \emph{ROS Commander} (ROSCo).
Generic parametrised skills are developed by experts in the respective field.
End-users can combine them by visual programming in the form of hierarchical finite
state machines (HFSM).
These state machines are a generic representation of a skill which then can
be deployed.
Other users can load these state machines and adapt the skill to their local environment,
e.g. by teaching where certain behaviours should be applied to.
Our system uses a mixed paradigm in which the generic skills are either created
by the end-user by visual programming and subsequent autonomous playing or by
field experts which make the corresponding controller available.

Alexandrova et al. designed a flow-based visual programming language called
\emph{RoboFlow} \cite{alexandrovaroboflow}.
Skills are either created top-down, i.e. by implementing a skill from scratch
by drag \& drop, or bottom-up, i.e. demonstrating the solution for one situation
and editing it afterwards in a graphical interface \cite{alexandrova2014robot}.
Our system basically follows the top-down approach, however, the user
can add parts of the skill by kinesthetic teaching.
These parts can be refined by policy reinforcement learning
\cite{Hangl-2015-ICAR, koberpower, theodorou2010generalized}.

Similarly, W\"achter et al. introduced the \emph{ArmarX} framework
which is provides an abstract view on robot programs following at a statechart concept.
Just as our framework, \emph{ArmarX} also follows the idea of an integrated software
suite based on C++ in which skill hierarchies can be created. 

All these works \cite{jackson2007microsoft, kim2007programming, nguyen2013ros,
alexandrovaroboflow, alexandrova2014robot} use visual programming for skills programming.
The key distinction of our approach is the integrated system
for autonomous skill acquisition.
In our system, the user can provide simple basic behaviours solving a
task in on specific situation.
The rest, i.e. generalisation to different situations, is then learned
autonomously by robotic playing.
Further, none of these systems provides an integrated autonomous testing framework.
\section{A skill-based programming paradigm}
In this section we sketch our skill acquisition paradigm based
on technology for developmental robotics.
We provide a formal definition of behaviours and skills by unifying
problem formulations from previous work
\cite{Hangl-2016-IROS, hangl2017tro, Hangl-2017-IROS},
c.f. section~\ref{sec:skilldef}, and give a high-level description of our
framework, c.f. section~\ref{sec:systemarchitecture}.
Sections~\ref{sec:autonomousplay} and \ref{sec:autonomoustesting} sketch
concrete solutions to the problem definitions given in this section.
\subsection{Behaviours and skills}
\label{sec:skilldef}
A \emph{behaviour} $b : S \rightarrow S, \, s \mapsto s'$ is a function that changes the
state $s \in S$ of an environment to some other state $s' \in S$.
The (partially unknown) states $s, \, s'$ consist of both, the external state of the
environment and the internal state of the robot.
Behaviours are basically everything a robot could potentially do, from low-level
behaviours such as simple point to point movements to high-level
behaviours such as grasping.
Behaviours do not come with a notion of a goal or task-specific success.
For this purpose we introduce a \emph{skill}
$\sigma = \left( b^{\sigma}, \, \text{Success}^{\sigma} \right)$, which comes with
a success predicate $\text{Success}^{\sigma}(s)$.
It defines whether or not the state $s \in S$ is a target state.
A skill execution is successful, if the behaviour $b^{\sigma}$ executed in state
$s \in S$ transforms the system such that
\begin{equation}
	\text{Success}^{\sigma}\left( b^{\sigma}(s) \right) = \text{true}
\end{equation}
The set
$D^{\sigma} = \{ s \, | \, \text{Success}^{\sigma}\left( b^{\sigma}(s) \right) = \text{true}  \}$
is called the \emph{domain of applicability} (DoA) of the skill $\sigma$.
In this paper, the domain of applicability is increased by preparing the
environment such that the basic behaviour can be executed to solve the task.
This is done by executing behaviour sequences of length $k$ with the property
\begin{equation}
	\text{Success}^{\sigma}\left(b_k \circ \dots \circ b_1 \circ b^{\sigma}(s') \right) = \text{true}
\end{equation}
such that $s' \in S$ was not yet in the DoA of the skill with $s' \notin D^{\sigma}$.
The success predicate can either be given from the outside, e.g. by a human teacher, or by
training a sensor model of successful executions.
\subsection{Sensor data and function call profiles}
\label{sec:sensordataandprofiles}
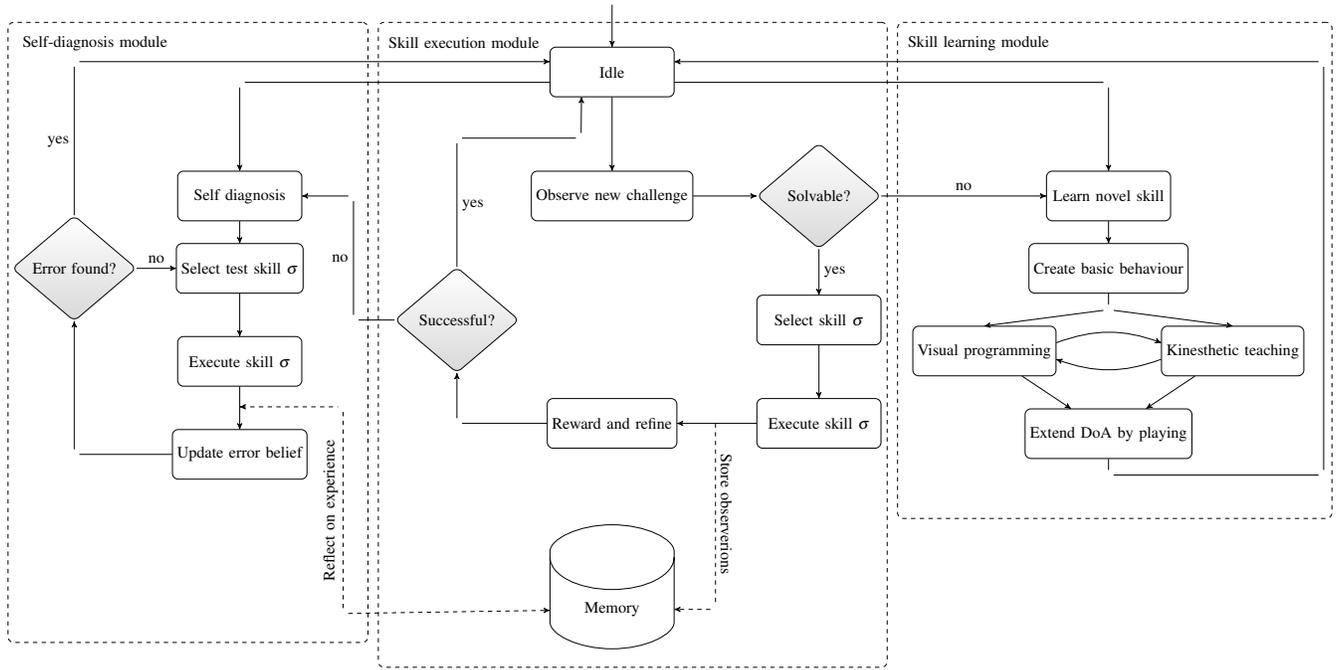
\begin{figure*}[t!]
\centering
  \scalebox{0.55} {
\begin{tikzpicture}[every node/.style = {shape=rectangle, rounded corners},>=stealth',bend angle=45, auto, font=\fontsize{11}{20}\selectfont,
]

  \tikzstyle{place}=[circle,thick,draw=blue!75,fill=blue!20,minimum size=6mm]

  \begin{scope}
  
	\draw node[rectangle, minimum width=8.7cm, minimum height=15.0cm, dashed, draw, xshift = -10.25cm, yshift = -6.3cm] (box1) {};
	\draw node[rectangle, minimum width=12.3cm, minimum height=15.6cm, dashed, draw, xshift = 9.75cm, yshift = -0.3cm, right of = box1] (box2) {};
	\draw node[rectangle, minimum width=10.5cm, minimum height=12.0cm, dashed, draw, xshift = 21.4cm, yshift = 1.5cm, right of = box1] (box3) {};
	
	\node[text width=5cm,align=left, left of = box1, yshift = 7.0cm, xshift = -0.5cm] (label1) {Self-diagnosis module};
	\node[text width=5cm,align=left, right of = label1, xshift = 7.85cm] (label2) {Skill execution module};
	\node[text width=5cm,align=left, right of = label2, xshift = 11.55cm] (label3) {Skill learning module};
  
	\draw node[rectangle, minimum width=3cm, minimum height=1.2cm,draw] (idle) {Idle};
	\node (memory) [cylinder, shape border rotate=90, draw, minimum height=3.0cm, minimum width=3cm, below of = idle, yshift = -12cm] {Memory};
	
	\draw node[rectangle, minimum width=3cm, minimum height=1.2cm,draw, below of = idle, yshift = -2cm] (exec1) {Observe new challenge};
	\draw node[rectangle, minimum width=3cm, minimum height=1.2cm,draw, left of = exec1, xshift = -8cm] (diag1) {Self diagnosis};
	\draw node[rectangle, minimum width=3cm, minimum height=1.2cm,draw, right of = exec1, xshift = 11cm] (learn1) {Learn novel skill};
	
	\draw node[decision, minimum width=3cm, minimum height=1.2cm,draw, right of = exec1, xshift = 4cm] (exec2) {Solvable?};
	\draw node[rectangle, minimum width=3cm, minimum height=1.2cm,draw, below of = exec2, yshift = -2cm] (exec3) {Select skill $\sigma$};
	\draw node[rectangle, minimum width=3cm, minimum height=1.2cm,draw, below of = exec3, yshift = -1.5cm] (exec4) {Execute skill $\sigma$};
	\draw node[rectangle, minimum width=3cm, minimum height=1.2cm,draw, left of = exec4, xshift = -4cm] (exec5) {Reward and refine};
	\draw node[decision, minimum width=3cm, minimum height=1.2cm,draw, left of = exec5, yshift = 2.5cm, xshift = -2.75cm] (exec6) {Successful?};
	
	\draw node[rectangle, minimum width=3cm, minimum height=1.2cm,draw, below of = diag1, yshift = -0.75cm] (diag2) {Select test skill $\sigma$};
	\draw node[rectangle, minimum width=3cm, minimum height=1.2cm,draw, below of = diag2, yshift = -1.25cm] (diag3) {Execute skill $\sigma$};
	\draw node[rectangle, minimum width=3cm, minimum height=1.2cm,draw, below of = diag3, yshift = -1.25cm] (diag4) {Update error belief};
	\draw node[decision, minimum width=3cm, minimum height=1.2cm,draw, left of = diag2, xshift = -3cm] (diag5) {Error found?};
	
	\draw node[rectangle, minimum width=3cm, minimum height=1.2cm,draw, below of = learn1, yshift = -0.75cm] (learn2) {Create basic behaviour};
	\draw node[rectangle, minimum width=3cm, minimum height=1.2cm,draw, below of = learn2, yshift = -1cm, xshift = -3cm] (learn3a) {Visual programming};
	\draw node[rectangle, minimum width=3cm, minimum height=1.2cm,draw, below of = learn2, yshift = -1cm, xshift = 3cm] (learn3b) {Kinesthetic teaching};
	\draw node[rectangle, minimum width=3cm, minimum height=1.2cm,draw, below of = learn3a, yshift = -1cm, xshift = 3cm] (learn4) {Extend DoA by playing};
	
	\path[->] ($(learn3a.east) + (0.0, 0.2)$) edge [line width=0.2mm, bend left=20] node {} ($(learn3b.west) + (0.0, 0.2)$);
	\path[->] ($(learn3b.west) + (0.0, -0.2)$) edge [line width=0.2mm, bend left=20] node {} ($(learn3a.east) + (0.0, -0.2)$);

	\node[dummy] (dummyentry) [above of = idle, yshift = 0.75cm] {};
	\path[->] (dummyentry) edge [line width=0.2mm] node {} (idle.north);
	
	\node[dummy] (dinit1) [above of = diag1, yshift = 1.75cm] {};
	\path[-] ($(idle.west) + (0, -0.25)$) edge [line width=0.2mm] node {} (dinit1);
	\path[->] (dinit1) edge [line width=0.2mm] node {} (diag1);
	
	\path[->] (idle.south) edge [line width=0.2mm] node {} (exec1.north);
	
	\node[dummy] (dinit3) [above of = learn1, yshift = 1.75cm] {};
	\path[->] (dinit3) edge [line width=0.2mm] node {} (learn1);
	\path[-] ($(idle.east) + (0, -0.25cm)$) edge [line width=0.2mm] node {} (dinit3);
	
	\path[->] (exec1.east) edge [line width=0.2mm] node {} (exec2.west);
	\path[->] (exec2.east) edge [line width=0.2mm] node {no} (learn1.west);
	\path[->] (exec2.south) edge [line width=0.2mm] node {yes} (exec3.north);
	\path[->] (exec3.south) edge [line width=0.2mm] node {} (exec4.north);
	\path[->] (exec4) edge [line width=0.2mm] node {} (exec5);
	
	\node[dummy] (d1) [left of = exec5, xshift = -2.75cm] {};
	\path[-] (exec5) edge [line width=0.2mm] node {} (d1);
	\path[->] (d1) edge [line width=0.2mm] node {} (exec6);
	
	\node[dummy] (d2) [left of = idle, xshift = -2.75cm, yshift = -1.6cm] {};
	\path[-] (exec6.north) edge [line width=0.2mm, right] node {yes} (d2);
	\node[dummy] (d2b) [right of = d2, xshift = 2cm] {};
	\path[-] (d2) edge [line width=0.2mm, right] node {} (d2b);
	\path[->] (d2b) edge [line width=0.2mm, right] node {} ($(idle.south) + (-0.74, 0)$);
	
	\node[dummy] (d3) [left of = exec6, xshift = -1.5cm] {};
	\path[-] (exec6) edge [line width=0.2mm] node {} (d3);
	\node[dummy] (d4) [right of = diag1, xshift = 1.75cm] {};
	\path[-] (d3) edge [line width=0.2mm] node {no} (d4);
	\path[->] (d4) edge [line width=0.2mm] node {} (diag1);
	
	\path[->] (diag1) edge [line width=0.2mm] node {} (diag2);
	\path[->] (diag2) edge [line width=0.2mm] node {} (diag3);
	\path[->] (diag3) edge [line width=0.2mm] node {} (diag4);

	\node[dummy] (diagdummy1) [left of = diag4, xshift = -3cm] {};
	\path[-] (diag4) edge [line width=0.2mm] node {} (diagdummy1);
	\path[->] (diagdummy1) edge [line width=0.2mm] node {} (diag5);
	
	\path[->] (diag5) edge [line width=0.2mm] node {no} (diag2);
	
	\node[dummy] (diagdummy2) [left of = idle, xshift = -12cm, yshift = 0.25cm] {};
	\path[-] (diag5.north) edge [line width=0.2mm] node {yes} (diagdummy2);
	\path[->] (diagdummy2) edge [line width=0.2mm] node {} ($(idle.west) + (0, 0.25)$);
	
	\path[->] (learn1) edge [line width=0.2mm] node {} (learn2);	
	
	\node[dummy] (dummylearning1) [below of = learn2] {};
	\path[-] (learn2) edge [line width=0.2mm] node {} (dummylearning1.north);
	\path[->] (dummylearning1) edge [line width=0.2mm] node {} (learn3a.north);
	\path[->] (dummylearning1) edge [line width=0.2mm] node {} (learn3b.north);
	
	\path[->] (learn3a) edge [line width=0.2mm] node {} (learn4);
	\path[->] (learn3b) edge [line width=0.2mm] node {} (learn4);
	
	\node[dummy] (dummylearning2) [below of = learn4] {};
	\path[-] (learn4) edge [line width=0.2mm] node {} (dummylearning2);
	\node[dummy] (dummylearning3) [right of = dummylearning2, xshift = 4.2cm] {};
	\path[-] (dummylearning2) edge [line width=0.2mm] node {} (dummylearning3);
	\node[dummy] (dummylearning4) [above of = dummylearning3, yshift = 9cm] {};
	\path[-] (dummylearning3) edge [line width=0.2mm] node {} (dummylearning4);
	\path[->] (dummylearning4) edge [line width=0.2mm] node {} ($(idle.east) + (0, 0.25cm)$);
	
	\node[dummy] (dummydiag5) [right of = diag4, yshift = 1.15cm, xshift = 1.5cm] {};
	\path[<-] ($(diag4.center) + (0, 1.15cm)$) edge [dashed, line width=0.2mm] node {} (dummydiag5);
	\node[dummy] (dummydiag6) [below of = dummydiag5, yshift = -4cm, xshift = -0.0cm] {};
	\path[-] (dummydiag5) edge [dashed, line width=0.2mm, above] node [rotate = 90] {Reflect on experience} (dummydiag6);
	\path[->] (dummydiag6) edge [dashed, line width=0.2mm] node {} (memory);
	
	\node[dummy] (dummymemory2) [left of = exec4, yshift = -4.5cm, xshift = -1.5cm] {};
	\path[-] ($(exec4.center) + (-2.5cm, 0)$) edge [dashed, line width=0.2mm, above] node [rotate = -90] {Store observerions} (dummymemory2);
	\path[->] (dummymemory2) edge [dashed, line width=0.2mm] node {} (memory);
	
  \end{scope}
\end{tikzpicture}
}
\caption{
System architecture illustrating the life-long learning scheme within our programming paradigm.
It consists of three major parts, i.e. the \emph{skill execution module}, the \emph{self-diagnosis module}
and the \emph{skill learning module}.
The memory is another core component of our system and stores data required by other components, e.g. sensor data or function call profiles.
}
\label{fig:systemarchitecture}
\end{figure*}

An essential part of the proposed programming paradigm is the storage
of experience gathered throughout the lifetime of the robot.
Whenever a skill or behaviour is executed, two matrices containing \emph{sensor data}
and \emph{function call profiles} are stored.
The sensor data per execution of the skill $\sigma$ with duration $T$ is given by
\begin{equation}
\mathbf{M}_{\sigma}(\mathbf{s}) =
\begin{blockarray}{cccc}
	\begin{block}{(ccc)c}
		\mathbf{m}_1(0) & \dots & \mathbf{m}_1(T) & \text{sensor } 1 \\
		\mathbf{m}_2(0) & \dots & \mathbf{m}_2(T) & \text{sensor } 2 \\
		\vdots & \ddots & \vdots & \downarrow \\
		\mathbf{m}_M(0) & \dots & \mathbf{m}_M(T) & \text{sensor } M  \\
	\end{block}
	t = 1 & \xrightarrow{\Delta t} & t = T & \\
\end{blockarray}
\end{equation}
The function call profile matrix stores information on how many instances
of all functions are active over the course of the execution
of a skill $\sigma$ and is given by
\begin{equation}
\mathbf{F}_{\sigma}(\mathbf{s}) =
\begin{blockarray}{cccc}
	\begin{block}{(ccc)c}
		fc_1(0) & \dots & fc_1(T) & \text{function } 1 \\
		fc_2(0) & \dots & fc_2(T) & \text{function } 2 \\
		\vdots & \ddots & \vdots & \downarrow \\
		fc_F(0) & \dots & fc_F(T) & \text{function } F  \\
	\end{block}
	t = 1 & \xrightarrow{\Delta t} & t = T & \\
\end{blockarray}
\label{equ:fingerprintmatrix}
\end{equation}
The database contains information on which skill requires certain
hardware configurations and both matrices are automatically stored to a
database when a skill is executed.
\subsection{System architecture}
\label{sec:systemarchitecture}
This section describes the key ingredients and their interplay in the context
of the proposed programming paradigm and is independent of concrete methods.
The system architecture is shown in Fig.~\ref{fig:systemarchitecture}.
Our life-long learning / programming approach consists of three major parts,
i.e. \emph{skill execution}, \emph{skill learning} and \emph{self-diagnosis}.
\subsubsection{Skill execution module}
\label{sec:shortskillexecution}
The skill execution module is responsible for executing well-trained skills.
The robot is confronted with a task and a decision on whether or not the task
can be solved with the available skill repertoire is made.
If a task cannot be solved, the skill learning module is called, which involves
external input from a programmer or teacher, c.f. section~\ref{sec:shortskilllearning}.
If the task is solvable, the appropriate skill is selected and executed by
performing a behaviour sequence $b_k \circ \dots \circ b_1 \circ b^{\sigma}(s)$
in the current state $s \in S$ which solves the task.
The execution is rewarded and the models are refined if the execution scheme
supports this.
During the execution of skills all experiences, i.e. sensor data and function call
profiles, are stored in the memory of the robot in order to use it for learning
and self-diagnosis.
\subsubsection{Skill learning module}
\label{sec:shortskilllearning}
The skill learning module can either be invoked out of idleness of the robot
or if a presented challenge cannot be solved by the robot.
The goal is to acquire a novel skill $\sigma$ with a large DoA.
In our setting, this requires two parts, i.e. teaching of the basic behaviour
$b^{\sigma}$ and identification of state dependent behaviour sequences
$b_k \circ \dots \circ b_1 \circ b^{\sigma}(s)$ which enlarge the DoA.
In this work, basic behaviours can be implemented by visual programming
(section~\ref{sec:graphicalprogramming}),
by kinesthetic teaching or a mixture of both.
In order to enlarge the DoA of the novel skill we propose to use autonomous
playing, c.f. section~\ref{sec:autonomousplay}.
\begin{figure*}[t!]
  \centering
  \subfloat[Schema of the episodic and compositional memory (ECM) for
  a certain skill $\sigma$.
  Skills are executed by performing a random walk through the ECM according to
  the probabilities that are assigned to the transitions.
  A skill execution consists of the sequence \emph{sensing action} $\rightarrow$
  \emph{perceptual state estimation} $\rightarrow$ \emph{environment preparation}
  $\rightarrow$ \emph{basic behaviour execution}.
  Skills are trained by using \emph{projective simulation}.
  ]{\label{fig:autonomousplaying}
  \resizebox{0.525\textwidth}{!}{%
  	\begin{tikzpicture}[every node/.style = {shape=rectangle, rounded corners},>=stealth',bend angle=45, auto, font=\fontsize{11}{20}\selectfont,
]

  \tikzstyle{place}=[circle,thick,draw=blue!75,fill=blue!20,minimum size=6mm]

  \begin{scope}
  
  \node[text width=5cm,align=left, yshift = -0.0cm, xshift = -5.7cm] (label1) {Layer 1};
  \node[text width=5cm,align=left, yshift = -2.0cm, xshift = -5.7cm] (label2) {Layer 2};
  \node[text width=5cm,align=left, yshift = -3.1cm, xshift = -5.7cm] (label3) {Layer 3};
  \node[text width=5cm,align=left, yshift = -5.6cm, xshift = -5.7cm] (label4) {Layer 4};
  \node[text width=5cm,align=left, yshift = -8.5cm, xshift = -5.7cm] (label5) {Layer 5};
  
	\draw node[rectangle, minimum width=3cm, minimum height=1.2cm,draw] (exec1) {Execute skill $\sigma$};
	\draw node[rectangle, minimum width=3cm, minimum height=1.2cm,draw, below of = exec1, xshift = -5cm, yshift = -1.0cm] (sensing1) {Sensing $se_1$};
	\draw node[rectangle, minimum width=3cm, minimum height=1.2cm,draw, below of = exec1, xshift = 0cm, yshift = -1.0cm] (sensing2) {Sensing $\dots$};
	\draw node[rectangle, minimum width=3cm, minimum height=1.2cm,draw, below of = exec1, xshift = 5cm, yshift = -1.0cm] (sensing3) {Sensing $se_H$};
	
	\draw node[rectangle, minimum width=5cm, minimum height=1.2cm,draw, below of = sensing1, xshift = 0cm, yshift = -1.0cm] (state1sum) {};	
	\draw node[circle, minimum width=1cm,draw, below of = sensing1, xshift = -1.5cm, yshift = -1.0cm] (state11) {$e^1_1$};
	\draw node[circle, minimum width=1cm,draw, below of = sensing1, xshift = 0cm, yshift = -1.0cm] (state12) {$\dots$};
	\draw node[circle, minimum width=1cm,draw, below of = sensing1, xshift = 1.5cm, yshift = -1.0cm] (state13) {$e^1_{I_1}$};
	
	\draw node[rectangle, minimum width=3cm, minimum height=1.2cm,draw, below of = sensing2, xshift = 0cm, yshift = -1.0cm] (state2) {Perceptual states of $\dots$};
	
	\draw node[rectangle, minimum width=3cm, minimum height=1.2cm,draw, below of = sensing3, xshift = 0cm, yshift = -1.0cm] (state3) {Perceptual states of $se_H$};
	
	\draw node[rectangle, minimum width=2.5cm, minimum height=1.2cm,draw, below of = state2, xshift = -7cm, yshift = -1.5cm] (beh1) {Behaviour $b_1$};
	\draw node[rectangle, minimum width=3cm, minimum height=1.2cm,draw, below of = state2, xshift = -3.5cm, yshift = -1.5cm] (beh2) {Behaviour $b_2$};
	\draw node[rectangle, minimum width=3cm, minimum height=1.2cm,draw, below of = state2, xshift = 0cm, yshift = -1.5cm] (beh3) {Behaviour $\dots$};
	\draw node[rectangle, minimum width=3cm, minimum height=1.2cm,draw, below of = state2, xshift = 3.5cm, yshift = -1.5cm] (beh4) {Behaviour $b_J$};
	\draw node[rectangle, minimum width=3cm, minimum height=1.2cm,draw, below of = state2, xshift = 7cm, yshift = -1.5cm] (beh5) {Behaviour $b_{\text{void}}$};
	
	\draw node[rectangle, minimum width=3cm, minimum height=1.2cm,draw, below of = beh3, xshift = 0cm, yshift = -1.0cm] (basic1) {Basic behaviour $b^{\sigma}$};
	
	\path[->] (exec1) edge [line width=0.2mm] node {} (sensing1.north);
	\path[->] (exec1) edge [line width=0.2mm] node {} (sensing2.north);
	\path[->] (exec1) edge [line width=0.2mm] node {} (sensing3.north);
	
	\path[->] (sensing1) edge [line width=0.2mm] node {} (state11.north);
	\path[->] (sensing1) edge [line width=0.2mm] node {} (state12.north);
	\path[->] (sensing1) edge [line width=0.2mm] node {} (state13.north);
	
	\path[->] (sensing2) edge [line width=0.2mm] node {} ($(state2.north) + (-0.5cm, 0)$);
	\path[->] (sensing2) edge [line width=0.2mm] node {} ($(state2.north) + (-1.5cm, 0)$);
	\path[->] (sensing2) edge [line width=0.2mm] node {} ($(state2.north) + (0.5cm, 0)$);
	\path[->] (sensing2) edge [line width=0.2mm] node {} ($(state2.north) + (1.5cm, 0)$);
	
	\path[->] (sensing3) edge [line width=0.2mm] node {} ($(state3.north) + (-0.5cm, 0)$);
	\path[->] (sensing3) edge [line width=0.2mm] node {} ($(state3.north) + (-1.5cm, 0)$);
	\path[->] (sensing3) edge [line width=0.2mm] node {} ($(state3.north) + (0.5cm, 0)$);
	\path[->] (sensing3) edge [line width=0.2mm] node {} ($(state3.north) + (1.5cm, 0)$);
	
	\path[->] (state11.south) edge [line width=0.2mm] node {} (beh1.north);
	\path[->] (state11.south) edge [line width=0.2mm] node {} (beh2.north);
	\path[->] (state11.south) edge [line width=0.2mm] node {} (beh3.north);
	\path[->] (state11.south) edge [line width=0.2mm] node {} (beh4.north);
	\path[->] (state11.south) edge [line width=0.2mm] node {} (beh5.north);
	
	\path[->] (state12.south) edge [line width=0.2mm] node {} (beh1.north);
	\path[->] (state12.south) edge [line width=0.2mm] node {} (beh2.north);
	\path[->] (state12.south) edge [line width=0.2mm] node {} (beh3.north);
	\path[->] (state12.south) edge [line width=0.2mm] node {} (beh4.north);
	\path[->] (state12.south) edge [line width=0.2mm] node {} (beh5.north);
	
	\path[->] (state13.south) edge [line width=0.2mm] node {} (beh1.north);
	\path[->] (state13.south) edge [line width=0.2mm] node {} (beh2.north);
	\path[->] (state13.south) edge [line width=0.2mm] node {} (beh3.north);
	\path[->] (state13.south) edge [line width=0.2mm] node {} (beh4.north);
	\path[->] (state13.south) edge [line width=0.2mm] node {} (beh5.north);
	
	\path[->] (state2.south) edge [line width=0.2mm] node {} (beh1.north);
	\path[->] (state2.south) edge [line width=0.2mm] node {} (beh2.north);
	\path[->] (state2.south) edge [line width=0.2mm] node {} (beh3.north);
	\path[->] (state2.south) edge [line width=0.2mm] node {} (beh4.north);
	\path[->] (state2.south) edge [line width=0.2mm] node {} (beh5.north);
	
	\path[->] (state3.south) edge [line width=0.2mm] node {} (beh1.north);
	\path[->] (state3.south) edge [line width=0.2mm] node {} (beh2.north);
	\path[->] (state3.south) edge [line width=0.2mm] node {} (beh3.north);
	\path[->] (state3.south) edge [line width=0.2mm] node {} (beh4.north);
	\path[->] (state3.south) edge [line width=0.2mm] node {} (beh5.north);
	
	\path[->] (beh1.south) edge [line width=0.2mm] node {} (basic1.north);
	\path[->] (beh2.south) edge [line width=0.2mm] node {} (basic1.north);
	\path[->] (beh3.south) edge [line width=0.2mm] node {} (basic1.north);
	\path[->] (beh4.south) edge [line width=0.2mm] node {} (basic1.north);
	\path[->] (beh5.south) edge [line width=0.2mm] node {} (basic1.north);
	
  \end{scope}
\end{tikzpicture}
  }
  } \quad
  \subfloat[Flow diagram of the used autonomous self-diagnosis approach.
  A \emph{measurement observation model} (MOM) for sensor data
  and a \emph{functional profiling fingerprint} (FPF) for function call profiles
  is trained for each skill from observations during successful executions.
  ]{\label{fig:selfdiagnosis}
  \resizebox{0.365\textwidth}{!}{%
  	\begin{tikzpicture}[every node/.style = {shape=rectangle, rounded corners},>=stealth',bend angle=45, auto, font=\fontsize{11}{20}\selectfont,
]

  \tikzstyle{place}=[circle,thick,draw=blue!75,fill=blue!20,minimum size=6mm]

  \begin{scope}
  
	\draw node[rectangle, minimum height=1.4cm, draw] (test1) {Skill $\sigma$ failed};
	
	\draw node[rectangle, text width=3cm, minimum height=1.4cm, draw, right of = test1, xshift = 3cm] (test2) {Estimate $t_{\text{fail}}$ by using the MOM};
	
	\draw node[rectangle, text width=4cm, minimum height=1.4cm, draw, right of = test2, xshift = 3.5cm] (test3) {Compare function calls until $t_{\text{fail}}$ with FPF};
	
	\draw node[rectangle, text width=4cm, minimum height=1.4cm, draw, below of = test3, yshift = -1.5cm] (test4) {Bug in running functions more likely};
	
	\draw node[rectangle, text width=4cm, minimum height=1.4cm, draw, below of = test4, yshift = -1.5cm] (test5) {Bayesian update of $p_{\text{blame}}(f = f_i)$};
	
	\draw node[rectangle, text width=4cm, minimum height=1.4cm, draw, below of = test5, yshift = -1.5cm] (test6) {Optimise expected information gain $\argmax_{\sigma'}{\text{ }I(\sigma', p_{\text{blame}}})$};
	
	\draw node[rectangle, text width=3cm, minimum height=1.4cm, draw, left of = test6, xshift = -3.5cm] (test7) {Execute skill $\sigma'$};
	
	\draw node[decision, draw, left of = test7, xshift = -3cm] (test8) {Failed?};
	
	\draw node[rectangle, text width=3.2cm, minimum height=1.4cm, draw, left of = test5, xshift = -3.5cm] (test9) {Bug in running functions less likely};
	
	\path[->] (test1) edge [line width=0.2mm] node {} (test2);
	\path[->] (test2) edge [line width=0.2mm] node {} (test3);
	\path[->] (test3) edge [line width=0.2mm] node {} (test4);
	\path[->] (test4) edge [line width=0.2mm] node {} (test5);
	\path[->] (test5) edge [line width=0.2mm] node {} (test6);
	\path[->] (test6) edge [line width=0.2mm] node {} (test7);
	\path[->] (test7) edge [line width=0.2mm] node {} (test8.east);
	
	\draw node[dummy, left of = test9, xshift = -3.0cm] (d1) {};
	\path[-] (test8.north) edge [line width=0.2mm] node {} (d1);
	\path[->] (d1) edge [line width=0.2mm] node {no} (test9);
	
	\draw node[dummy, left of = test4, xshift = -7.5cm] (d2) {};
	\path[-] (d1) edge [line width=0.2mm] node {} (d2);
	\draw node[dummy, below of = test2, yshift = -1.5cm] (d3) {};
	\path[-] (d2) edge [line width=0.2mm] node {yes} (d3);
	\path[->] (d3) edge [line width=0.2mm] node {} (test2);
	
	\path[->] (test9) edge [line width=0.2mm] node {} (test5);
	
  \end{scope}
\end{tikzpicture}
  }
  }
\caption{Conceptual design on the used implementations for the \emph{skill execution module}
(Fig.~\ref{fig:autonomousplaying}) and the \emph{skill learning module}
(Fig.~\ref{fig:selfdiagnosis}).
The overall architecture is shown in Fig.~\ref{fig:systemarchitecture}.
}
\label{fig:prevwork}
\end{figure*}
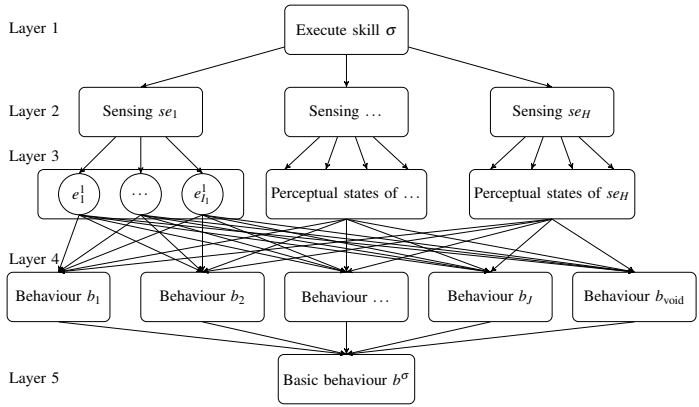
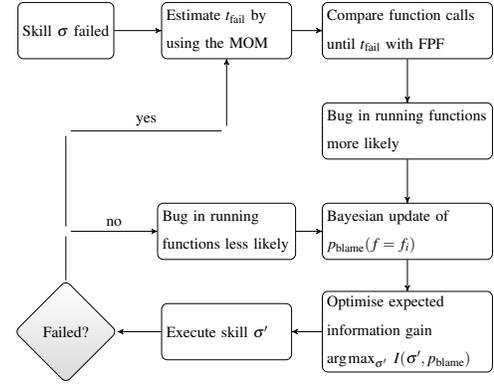
\subsubsection{Self-diagnosis module}
\label{sec:shortselfdiagnosis}
We emphasise the need of a way to autonomously identify bugs in the software
as part of the programming paradigm.
The self-diagnosis module is called if the robot is idle or if typically
successful skills start to fail.
Due to the life-long programming approach, the robot has access to a large
database of experiences, i.e. sensor data and function call profiles.
Skills are executed as test cases in order to identify functions and behaviours
which cause a problem that was not present before certain code changes.
The system returns a distribution $p_{\text{blame}}(f = f_i)$ defining
the probability of the function $f_i$ causing the problem.
In this paper we use the testing approach described in
section~\ref{sec:autonomoustesting}.
\subsubsection{Memory}
The memory is a core component of our architecture following the philosophy
of storing all available experiences made over the life-time of the robot.
Previous work has demonstrated a way on how to store sensor data to the non-SQL
database MongoDB based on listening on ROS topics and using this information
for fault analysis \cite{databasefaultdetection}.
We chose a different track in which hardware interfaces are provided within
our software that require developers of the control classes to provide a SQL
schema and \emph{store} functions for storing a current snapshot of the sensor data.
For all hardware used for the execution of a skill, the \emph{store} function
is called regularly.
Further, for all framework functions the start and end time of the execution
is stored in the database.
\section{Skill acquisition by autonomous playing}
\label{sec:autonomousplay}
The previous sections are agnostic of the concrete skill acquisition and self-diagnosis method.
In this section we provide a brief description of the skill learning method used in our prototype.
The reader may refer to previous work for more in-depth treatment \cite{hangl2017tro, Hangl-2016-IROS}.
The idea is teach a novel skill $\sigma$ by providing a basic behaviour $b^{\sigma}$
and to extend the DoA by learning how to prepare a situation in which the basic behaviour
can be applied without changes.
The robot tries out different combinations of preparatory behaviours in different situations.
We refer to this as autonomous playing.
The playing consists of two stages - in the first stage so-called \emph{perceptual states}
$e \in E^{se}$ are trained from interacting with the object by using sensing actions $se \in SE$.
Perceptual states are discrete concepts derived
from continuous sensor observations and capture the task-relevant information.
This might require different sensing actions for different tasks, e.g. poking on top of
a box can be used to identify whether or not a box is open, whereas sliding along the surface
can be used to determine the orientation of a book.
Different situations are prepared either by the robot or by a human teacher
and the robot creates a haptic database of sensor data observed in
different perceptual states.
A SVM is trained from the sensor data for perceptual state classification.

After this first playing phase a so-called \emph{episodic and compositional memory} (ECM)
of the form shown in Fig.~\ref{fig:autonomousplaying} is constructed.
The ECM is the basis of \emph{projective simulation} \cite{Briegel2012}
and consists of \emph{clips} that are connected by transitions with certain transition probabilities.
A skill is executed by a \emph{random walk} through the ECM and by executing the actions
along the path.
The transition probabilities along a rewarded path are increased, i.e. if the success predicate
of the skill $\sigma$ is fulfilled.
This forces the robot
to learn (i) which sensing action provides the best perceptual states; (ii) how to prepare
the environment correctly in a certain perceptual state.
When a skill is well-trained, i.e. the average success rate during playing is high, it
can be used as preparatory behaviour for other skills to train in the future.
This enables to robot to create skill hierarchies and to learn how to solve increasingly
complex problems.
The described basic version was extended by creative planning and active learning mechanisms
in followup work \cite{hangl2017tro}.
While playing, a simple environment model is trained which is used to
suggest potentially useful preparatory behaviours to the model-free playing system.
An in-depth treatment is provided in previous work \cite{Hangl-2016-IROS, hangl2017tro}.
\section{Autonomous skill-centric testing}
\label{sec:autonomoustesting}
For \emph{self-diagnosis module} we use a method for autonomous testing
developed in previous work \cite{Hangl-2017-IROS}.
During a skill execution, an observation
$o = \left(\mathbf{M}_{\sigma}(s), \, \mathbf{F}_{\sigma}(s) \right)$ is made.
From the matrices $\mathbf{M}_{\sigma}(s)$ (sensor data) and
$\mathbf{F}_{\sigma}(s)$ (function call profiles)
described in section~\ref{sec:sensordataandprofiles} the measurement
observation model (MOM) $p_{\sigma}(succ \, | \, \mathbf{M}, t)$ and the
functional profiling fingerprint (FPF)
$p_{\sigma}^f (fc \, | \, \mathbf{M}, \, t, \, succ = \text{true})$ for each
software function $f$ are trained.
The variables $s \in S$, $succ$, $t$ and $fc$ denote the environment state,
the execution success, the time step
and the number of running instances of the function $f$ respectively.
The MOM and the FPF are trained by an encoder / decoder neural network and by
a simple multivariate Gaussian model respectively.
A sketch of the testing scheme is shown in Fig.~\ref{fig:selfdiagnosis}.
If a skill $\sigma$ failed, the MOM is used to identify the time $t_{\text{fail}}$
at which the error occurred.
All functions that were running until this point are at least suspicious proportional
to the time distance to $t_{\text{fail}}$, and even more so if the matrix
$\mathbf{F}_{\sigma}(\mathbf{s})$ for a certain function strongly deviates from
the trained FPF.
If a function is running during a successful execution, it becomes less likely
that this function has a problem.
From this information the likelihood function $p_{\sigma}(o \, | \, f = f_i)$
is computed, and a Bayesian belief update is performed with
\begin{equation}
p_{\text{blame}}(f \, | o_{1 : T}, o, \sigma_{1 : T}, \sigma) \propto
	p_{\sigma}(o \, | \, f) p_{\text{blame}}(f \, | o_{1 : T}, \sigma_{1 : T})
\end{equation}

In a further step, our system selects the next skill $\sigma'$ to execute by
maximising the expected information gain $\mathbf{E}[I(\sigma')]$ about
which function causes problems.
The resulting distribution $p_{\text{blame}}(f = f_i)$ is estimated autonomously.
The developer can then either fix the bug if the knowledge is available or report
it to the responsible designer of the respective components.
\section{System integration}
In following section we describe practical considerations on essential key components
in order to achieve two major goals: the \emph{integration of skill acquisition
and the autonomous testing} and the \emph{design of the visual programming framework}.
\subsection{The \emph{kukadu} framework}
All the described methods are implemented in \emph{kukadu} - a framework
for autonomous robotics.
\emph{kukadu} provides several modules required in robotics:
\begin{itemize}
	\item Hardware control: Provides interfaces for hardware components like arms
	or depth-image cameras.
	Implementations for certain hardware are included.
	\item Path planning: Provides interfaces for path planning (joint space and Cartesian space)
	and kinematics.
	Bridges to path planning frameworks such as
	\emph{MoveIt}\footnote{\url{http://moveit.ros.org/}} or \emph{Komo} \cite{14-toussaint-KOMO}
	are available.
	\item Control policies: Provides interfaces for parametrised control policies 
	and standard methods such as DMPs (joint space \cite{schaal2006dynamic}
	and Cartesian space \cite{ude2014orientation}) are implemented.
	Policies can be trained by Kinesthetic teaching.
	\item Machine learning: Different standard machine learning techniques such as linear regression,
	Gaussian process regression, Support vector machines \cite{libsvm},
	unsupervised clustering are available.
	Certain policy reinforcement learning approaches like PI$^2$ \cite{theodorou2010generalized}
	or PoWER \cite{koberpower} and gradient descent methods are available.
	The algorithms are connected to the control policy interfaces.
	\item Computer vision: Interfaces for object localisation and pose estimation are defined.
	Elementary object localisation, integrated with the depth-image interfaces,
	based on fitting boxes to segmented point clouds is provided.
	\item Autonomous robotics: The approaches described in sections \ref{sec:autonomousplay}
	and \ref{sec:autonomoustesting} are implemented and connected to the
	available hardware interfaces.
	The methods are based on a skill interface which provides automatic controller
	generation or automatic storage of sensor snapshots.
\end{itemize}
We further provide a visual programming interface for
unexperienced developers and programmers without C++ knowledge.
The following sections explain key concepts on an abstract level that make this
possible inside \emph{kukadu}.
Detailed information can be found in the framework documentation.
\subsection{SQL-based robot memory}
Many ROS-based applications use the integrated MongoDB in order to store data.
This approach avoids the need for designing an SQL schema for robotics applications.
We chose a different approach by designing an extensible SQL schema, considering typical
data provided by hardware that is interfaced in the framework.
Certain hardware provides more information, e.g. new robotic arms might provide more
or different sensor data, which is why implementations of hardware interfaces can provide
their own additional schema in form of SQL files, which are installed on program startup.
Similar concepts hold for skill controllers, which are installed automatically, e.g.
storing the required hardware configurations.
More involved skill controllers can further install custom SQL
schema and store additional information.
\subsection{Factories}
In order to provide easy drag and drop functionality for end-users, it must be easy
to instantiate hardware control and skill objects without having to deal with constructors.
This is ensured by hardware and skill factories, which query the code on how to generate
instances from a code database provided by the respective developers of the interfaces.
Custom code for loading required properties from the database can also be defined.
Hardware can be created by providing a concrete instance name, e.g. \emph{left\_arm}.
Skills can be created by defining the skill name and the desired hardware configuration,
e.g. the \emph{simple grasping} skill in Fig.~\ref{fig:examplesimplegrasp} requires
a hardware configuration consisting of \emph{left\_arm, camera} and \emph{left\_hand}.
Further, the factory makes sure that only one controller instance exists at the
same time per hardware component.
\subsection{Skill interface and data acquisition}
A skill is generated by implementing an abstract skill controller class.
An instance of the respective controller is created by the skill factory using the
information stored in the database.
If a skill is executed, pre-implemented functions query all the required hardware.
A monitor regularly requests the used hardware to store snapshots of the
available sensor data to the database.
This is especially important for the generation of the skill-specific FPF and MOM,
c.f. sections~\ref{sec:sensordataandprofiles} and \ref{sec:autonomoustesting}.
The sensor data is connected to the respective starting and end times of concrete executions
and to the success information.
\subsection{Visual robot programming}
\label{sec:graphicalprogramming}
The visual programming tool enables the user to create simple robot programs
based on drag and drop.
We base our implementation on the \emph{cake}
framework\footnote{\url{https://github.com/cra16/cake-core}},
which generates C++ code from the graphic input.
In principle arbitrarily complex programs can be generated and
typical language constructs like loops or functions are available out of the box.
We further added custom blocks for robotic programming, namely \emph{skill} blocks
and \emph{hardware} blocks.
The programmer can drop multiple hardware components onto the canvas and the
interface suggests skills that are available for a certain hardware setting,
e.g. by using the arm, the hand and the camera the \emph{grasping} skill can be used.

We provide an initial set of different simple skills ranging from simple point to point
movement (Ptp) skills to more high-level skills like pushing.
Some skills require additional input which can be defined as well,
e.g. the \emph{JointPtp} skill needs the desired goal position.
\emph{kukadu} is implemented in C++ which does not come with a sophisticated
reflection API such as
\emph{Java}\footnote{\url{https://docs.oracle.com/javase/tutorial/reflect/}}.
Therefore, the supported properties of certain skills are extracted by generating
an XML-based representation of all member functions of the skill implementation classes
by using \emph{Doxygen}\footnote{\url{http://www.stack.nl/~dimitri/doxygen/}}.
\subsubsection{Skill implementation}
The developer can generate the code for a novel skill controller and test
it automatically on the robot.
We support an interactive programming paradigm in which programming can be
done directly in a certain environment by a mix of visual programming and
kinesthetic teaching.
In the task of placing a book to a shelf the developer can use the \emph{book grasping}
skill to grasp the book.
The \emph{book placement} skill can be tested up until this point which leaves
the robot in a state where the book is in the robot's hand.
The user can continue by showing how to place the book in the shelf
by kinesthetic teaching.
This will work only if the book in the correct pose, c.f. section~\ref{sec:autonomousplay}.
Instead of asking the user to consider even more cases, one can create a
basic behaviour out of the visual program and call the module for autonomous playing.
The correct treatment of different book orientations, i.e. the extension of the DoA
of the \emph{book placement} skill will then be learned by the robot.

If the user is happy with the result the novel skill can be installed and used
for the implementation of other skills as well.
This way more and more complex skills can be implemented and skill hierarchies
can be created.
\subsubsection{Skill testing}
During training and the usage of skills a large amount of sensor data is
automatically stored in the database.
If code is changed this could influence the implemented skills, e.g. if the
skill for moving the arm in Cartesian space starts using a different planner.
Such errors can be identified by the skill testing module \cite{Hangl-2017-IROS}
which is integrated as a separate tool.
The information on potentially problematic functions can be used by the developer
to consult experts on the respective components or to change used parameters.
This strongly reduces the need for the user to be familiar with
certain components in detail.
\section{Evaluation}
The separate components of our approach to robot programming, namely the
\emph{autonomous playing} \cite{Hangl-2016-IROS, hangl2017tro} and the
\emph{skill centric testing} \cite{Hangl-2017-IROS}, were evaluated
in their respective publications.
In this paper we investigate the applicability of the complete
and integrated system based on our novel robot programming paradigm.
We demonstrate that a wide variety of skills can be implemented with
our framework for visual robot programming.
In order to evaluate the skill acquisition,
we implemented basic behaviours for different skills by visual programming.
The DoA is then extended by autonomous playing.
\begin{table}[h]
\centering
\begin{tabular}{ l l l l }
Name & Description & Type & Visual \\
\hline\\
\pbox{1.5cm}{Sliding} & \pbox{3cm}{Slides along the object's surface} & Se & Yes \\[0.2cm]
\pbox{1.5cm}{Poking} & \pbox{3cm}{Pokes on top of the object} & Se & Yes \\[0.2cm]
\pbox{1.5cm}{Pressing} & \pbox{3cm}{Presses the object between the hands} & Se & Yes \\[0.2cm]
\pbox{1.5cm}{Joint movement} & \pbox{3cm}{Moves joints to specified position} & B & No \\[0.2cm]
\pbox{1.5cm}{Cartesian movement} & \pbox{3cm}{Moves end-effector to Cartesian position} & B & No \\[0.2cm]
\pbox{1.5cm}{Move home} & \pbox{3cm}{Move to initial position} & B & Yes \\[0.2cm]
\pbox{1.5cm}{Close hand} & \pbox{3cm}{Close the hand} & B & Yes \\[0.2cm]
\pbox{1.5cm}{Open hand} & \pbox{3cm}{Open the hand} & B & Yes \\[0.2cm]
\pbox{1.5cm}{Change stiffness} & \pbox{3cm}{Change impedance settings of the arm} & B & No \\[0.2cm]
\pbox{1.5cm}{Push to body} & \pbox{3cm}{Push an object towards the body} & B & Yes \\[0.2cm]
\pbox{1.5cm}{Push from body} & \pbox{3cm}{Push object away from the body} & B & Yes \\[0.2cm]
\pbox{1.5cm}{Push to position} & \pbox{3cm}{Push object to a certain position} & B & Yes \\[0.2cm]
\pbox{1.5cm}{Push to orientation} & \pbox{3cm}{Rotate the object to a certain orientation} & B & No \\[0.2cm]
\pbox{1.5cm}{Simple grasp} & \pbox{3cm}{Place end-effector on top of an object and close the hand} & B & Yes \\[0.2cm]
\pbox{1.5cm}{Pick and place} & \pbox{3cm}{Grasp an object and place it in a box} & B & Yes \\[0.2cm]
\pbox{1.5cm}{Press button} & \pbox{3cm}{Presses a button at a fixed position} & B & Yes \\[0.2cm]
\pbox{1.5cm}{Book grasping} & \pbox{3cm}{Grasp a book from a rigid table} & Sk & Yes \\[0.2cm]
\pbox{1.5cm}{Shelf placement} & \pbox{3cm}{Place an object in a shelf} & Sk & Yes \\[0.2cm]
\pbox{1.5cm}{Tower disassembly} & \pbox{3cm}{Disassembles a tower of boxes} & Sk & Yes \\[0.2cm]
\pbox{1.5cm}{Localise object} & \pbox{3cm}{Searches for a defined object on the table} & B & No \\[0.2cm]
\hline
\\
\end{tabular}
\caption{List of the implemented skills and behaviours by using the \emph{kukadu} framework.
The \emph{Visual} column indicates whether or not the corresponding controller was
implemented by visual programming.
The \emph{Type} corresponds to the controller type (\emph{Se} $\equiv$ sensing action,
\emph{B} $\equiv$ behaviour, \emph{Sk} $\equiv$ skill trained with autonomous playing).}
\label{tab:implementedskills}
\end{table}
\begin{figure*}[t!]
\centering
\subfloat[
Visual program of a \emph{simple grasping} skill.
The robot moves to its' home position, detects some object on the table,
moves the end-effector to that position and closes the hand.
]{
	\label{fig:examplesimplegrasp}
	\includegraphics[width=0.58\textwidth]{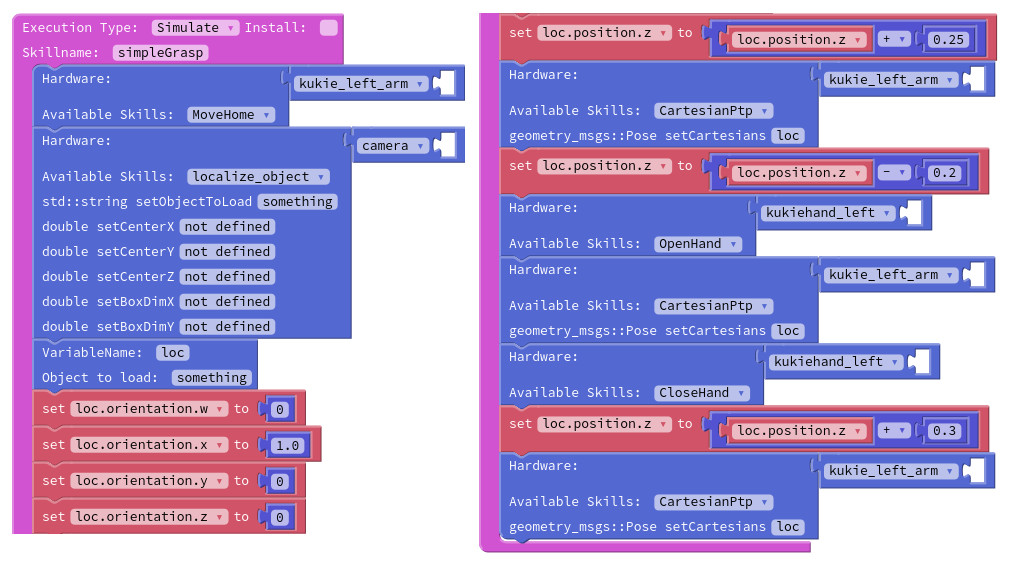}
} \quad
\subfloat[
Visual program of a \emph{pick and place} skill which reuses the simple
grasping skill in Fig.~\ref{fig:examplesimplegrasp}.
]{
	\label{fig:examplepickandplace}
	\includegraphics[width=0.36\textwidth]{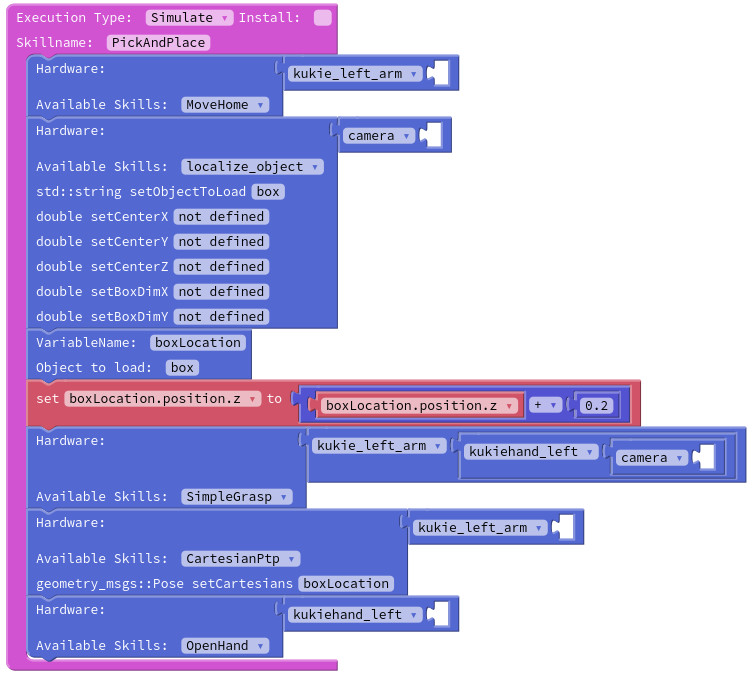}
}
\caption{Exemplary visual programs of a \emph{simple grasping} skill and a \emph{pick and
place} skill.
The creation of skill hierarchies, e.g. the \emph{pick and place} skill uses the
\emph{grasping} skill, is demonstrated.
}
\label{fig:examples}
\end{figure*}
\begin{figure}
\centering
\vspace{0.2cm}
\includegraphics[width=0.5\textwidth]{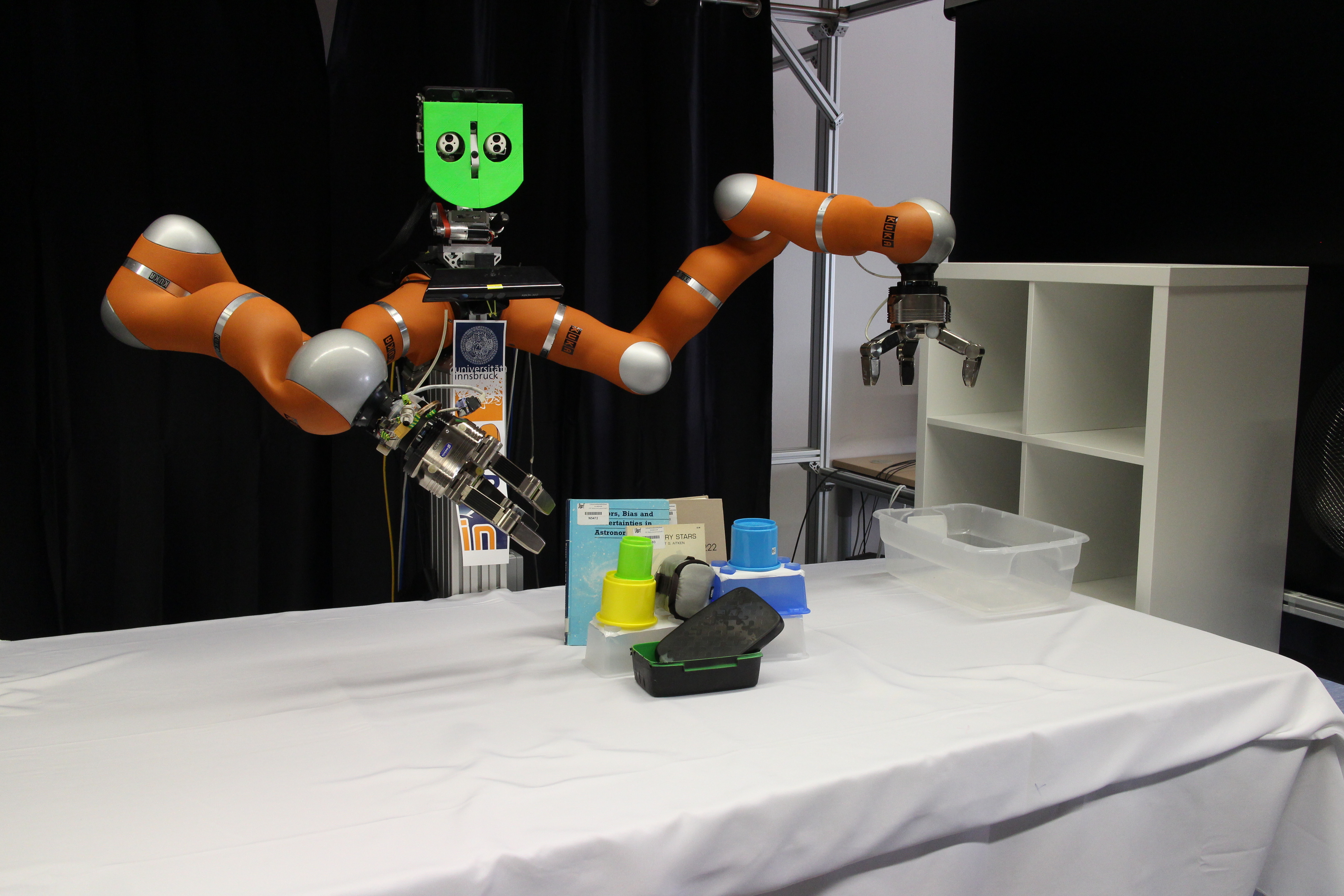}
\caption{The robot setting (2 KUKA LWR 4+ with attached Schunk SDH
grippers) and the used objects.}
\label{fig:robotsetting}
\end{figure}
\subsection{Robot setting}
The used robot setting is shown in Fig.~\ref{fig:robotsetting}.
It consists of two KUKA LWR 4+ robotic arms mounted on a metal pillar.
A Schunk SDH gripper is mounted to each one of the robot's arms.
A Kinect camera is mounted above the robot for
object recognition and localisation.
In order to develop and test the implemented skills several objects
from the YCB dataset were used \cite{calli2015ycb}.
We further used other objects such as books of different sizes
and Ikea furniture and boxes.
\subsection{Implemented skills}
We demonstrate the generality of our approach by implementing a wide range
of different skills and behaviours by visual programming and autonomous playing.
A list of these skills including a short description can be found in
Table~\ref{tab:implementedskills}.
\subsubsection{Conventional behaviour programming}
Fig.~\ref{fig:examplesimplegrasp} shows a \emph{simple grasping} skill in which
the object is localised and the end-effector is moved on top of it.
The hand is closed and the object is lifted.
The skill can directly be tested and installed if the user is satisfied with the result.
It is immediately available for further use, e.g. with the
\emph{pick and place} skill shown in Fig.~\ref{fig:examplepickandplace}.
This way skill hierarchies can be created.

We further implemented a \emph{push to position} behaviour which demonstrates
how more complex control structures such as loops can be used.
The end-effector follows a Cartesian path from the object's current position
to the desired position.
The path is planned in Cartesian space in a loop with intermediate
milestone positions.
\subsubsection{Implementation of skills by autonomous playing}
Up until now none of the described behaviours required autonomous playing.
We implemented a \emph{book grasping} skill and a \emph{tower disassembly} skill.
The basic behaviour for \emph{book grasping} is programmed according to the
description in section~\ref{sec:autonomousplay}, c.f.
video\footnote{\url{https://iis.uibk.ac.at/public/shangl/iros2016/iros.mpg}}.
The book is localised by using the \emph{localise object} behaviour
and is pushed to a position in front of the robot with the
\emph{push to position} behaviour.
The basic behaviour can be executed up until this point and the user
can use kinesthetic teaching in order to push the book towards the robot
until it is squeezed between both hands.
The system automatically learns a DMP out of the demonstrated trajectory and makes
it available for further use during visual programming.
The user can then continue to implement the rest of the basic behaviour.

This approach only works if the book is rotated correctly such that it can
be lifted at the spine.
The robot learned how to deal with different rotations by autonomous playing.
It automatically identified that the \emph{sliding} behaviour is best
suited to estimate the book's orientation and that pushing the book
to the correct rotation yields success.

The skill of \emph{disassembling a tower of boxes} skill shows that the basic behaviour can
even be left empty, c.f.
video\footnote{\url{https://iis.uibk.ac.at/public/shangl/iros2016/iros.mpg}}.
This way the robot learns how to coordinate already existing behaviours and figures
out that the tower can be disassembled by performing $h$ consecutive
\emph{pick and place} actions, where $h$ is the height of the tower.
The height can be estimated by \emph{poking} on top of the tower.

Some skills were not programmable with visual programming due to the restrictions
in order to provide a simple programming framework.
This can be the case if complex sensor data processing or hardware functions are required.
An example is the \emph{rotation by pushing} skill which requires fine-grained direct
control of the shape of the trajectory.
Another example is the \emph{change stiffness} skill which changes the stiffness settings
of the robot's impedance mode.
In such cases the skills can be implemented by experts and can be made available to
unexperienced programmers for usage.
\subsection{Testing}
The testing functionality can be called as a separate tool in case the robot
either keeps failing on a skill that was well-trained before or if the user
triggers it manually.
The testing framework is especially helpful in case the user changes the implementation
of a skill or behaviour, e.g. if properties of used behaviours are changed or if
the structure of the implemented behaviour is changed completely.
An example could be that the used planner for executing a Cartesian
plan was switched.
Suggestions of functions that may have caused the problem are then made.
These can be fixed by the user or forwarded to experts in the respective fields.
In prior work we showed that such errors can be found autonomously \cite{Hangl-2017-IROS}.
\section{Conclusion and Outlook}
In this work we presented a novel paradigm for robot programming which
adopts technology developed for developmental robotics in order to ease
the task for robot programmers.
The programmer only has to implement a basic behaviour that solves a
task for only one situation by interactive programming with visual programming
and kinesthetic teaching.
The robot then uses this program as a starting point in order to
extend the domain of applicability.

Further, the robot gathers a lot of sensor data during the execution of skills
which can then be used to identify bugs in the software if a skill stops working.

We evaluated our approach by implementing several behaviours and skills
with our new tool for visual programming and extended the DoA
by autonomous playing.

There are many ways along which our approach can be extended.
If one looks at the generated programs, many of the commands can be represented
as a simple sentence with a \emph{subject}, e.g. camera,
a \emph{predicate}, e.g. localise, and an \emph{object}, e.g. book.
This could be exploited to implement a verbal interface which could be
used in combination with visual programming.
Further, we plan to extend our developmental approaches in order to enable the robot
to learn the extension of the DoA more efficiently, e.g. by transferring knowledge
between skills.
Another interesting direction is to transfer skills between different robots, which
in general would enable robot programmers to implement skills which can easily
be shared with other robot owners.



\balance

\bibliographystyle{IEEEtran}
\bibliography{bibfile}

\begin{thebibliography}{10}
\providecommand{\url}[1]{#1}
\csname url@rmstyle\endcsname
\providecommand{\newblock}{\relax}
\providecommand{\bibinfo}[2]{#2}
\providecommand\BIBentrySTDinterwordspacing{\spaceskip=0pt\relax}
\providecommand\BIBentryALTinterwordstretchfactor{4}
\providecommand\BIBentryALTinterwordspacing{\spaceskip=\fontdimen2\font plus
\BIBentryALTinterwordstretchfactor\fontdimen3\font minus
  \fontdimen4\font\relax}
\providecommand\BIBforeignlanguage[2]{{%
\expandafter\ifx\csname l@#1\endcsname\relax
\typeout{** WARNING: IEEEtran.bst: No hyphenation pattern has been}%
\typeout{** loaded for the language `#1'. Using the pattern for}%
\typeout{** the default language instead.}%
\else
\language=\csname l@#1\endcsname
\fi
#2}}

\bibitem{hangl2017tro}
S.~Hangl, V.~Dunjko, H.~J. Briegel, and J.~Piater, ``Skill learning by
  autonomous robotic playing using active learning and creativity,'' e-Print
  arXiv:1706.08560, 2017.

\bibitem{Hangl-2016-IROS}
S.~Hangl, E.~Ugur, S.~Szedmak, and J.~Piater, ``{Robotic playing for
  hierarchical complex skill learning},'' in \emph{{IEEE/RSJ International
  Conference on Intelligent Robots and Systems}}, 2016.

\bibitem{Hangl-2017-IROS}
S.~Hangl, S.~Stabinger, and J.~Piater, ``{Autonomous Skill-centric Testing
  using Deep Learning},'' in \emph{{IEEE/RSJ International Conference on
  Intelligent Robots and Systems}}, 2017.

\bibitem{lieberman2006end}
H.~Lieberman, F.~Patern{\`o}, M.~Klann, and V.~Wulf, ``End-user development: An
  emerging paradigm,'' in \emph{End user development}.\hskip 1em plus 0.5em
  minus 0.4em\relax Springer, 2006, pp. 1--8.

\bibitem{ko2004six}
A.~J. Ko, B.~A. Myers, and H.~H. Aung, ``Six learning barriers in end-user
  programming systems,'' in \emph{Visual Languages and Human Centric Computing,
  2004 IEEE Symposium on}.\hskip 1em plus 0.5em minus 0.4em\relax IEEE, 2004,
  pp. 199--206.

\bibitem{resnick2009scratch}
M.~Resnick, J.~Maloney, A.~Monroy-Hern{\'a}ndez, N.~Rusk, E.~Eastmond,
  K.~Brennan, A.~Millner, E.~Rosenbaum, J.~Silver, B.~Silverman, \emph{et~al.},
  ``Scratch: programming for all,'' \emph{Communications of the ACM}, vol.~52,
  no.~11, pp. 60--67, 2009.

\bibitem{kim2007programming}
S.~H. Kim and J.~W. Jeon, ``Programming lego mindstorms nxt with visual
  programming,'' in \emph{Control, Automation and Systems, 2007. ICCAS'07.
  International Conference on}.\hskip 1em plus 0.5em minus 0.4em\relax IEEE,
  2007, pp. 2468--2472.

\bibitem{jackson2007microsoft}
J.~Jackson, ``Microsoft robotics studio: A technical introduction,'' \emph{IEEE
  Robotics \& Automation Magazine}, vol.~14, no.~4, 2007.

\bibitem{nguyen2013ros}
H.~Nguyen, M.~Ciocarlie, K.~Hsiao, and C.~C. Kemp, ``Ros commander (rosco):
  Behavior creation for home robots,'' in \emph{Robotics and Automation (ICRA),
  2013 IEEE International Conference on}.\hskip 1em plus 0.5em minus
  0.4em\relax IEEE, 2013, pp. 467--474.

\bibitem{riedo2013thymio}
F.~Riedo, M.~Chevalier, S.~Magnenat, and F.~Mondada, ``Thymio ii, a robot that
  grows wiser with children,'' in \emph{Advanced Robotics and its Social
  Impacts (ARSO), 2013 IEEE Workshop on}.\hskip 1em plus 0.5em minus
  0.4em\relax IEEE, 2013, pp. 187--193.

\bibitem{alexandrovaroboflow}
S.~Alexandrova, Z.~Tatlock, and M.~Cakmak, ``Roboflow: A flow-based visual
  programming language for mobile manipulation tasks,'' in \emph{2015 IEEE
  International Conference on Robotics and Automation (ICRA)}, May 2015, pp.
  5537--5544.

\bibitem{alexandrova2014robot}
S.~Alexandrova, M.~Cakmak, K.~Hsiao, and L.~Takayama, ``Robot programming by
  demonstration with interactive action visualizations.'' in \emph{Robotics:
  science and systems}, 2014.

\bibitem{Hangl-2015-ICAR}
S.~Hangl, E.~Ugur, S.~Szedmak, A.~Ude, and J.~Piater, ``{Reactive,
  Task-specific Object Manipulation by Metric Reinforcement Learning},'' in
  \emph{{17th International Conference on Advanced Robotics}}, 7 2015.

\bibitem{koberpower}
J.~Kober and J.~R. Peters, ``Policy search for motor primitives in robotics,''
  in \emph{Advances in Neural Information Processing Systems 21}, D.~Koller,
  D.~Schuurmans, Y.~Bengio, and L.~Bottou, Eds.\hskip 1em plus 0.5em minus
  0.4em\relax Curran Associates, Inc., 2009, pp. 849--856.

\bibitem{theodorou2010generalized}
E.~Theodorou, J.~Buchli, and S.~Schaal, ``A generalized path integral control
  approach to reinforcement learning,'' \emph{Journal of Machine Learning
  Research}, vol.~11, no. Nov, pp. 3137--3181, 2010.

\bibitem{databasefaultdetection}
T.~Niemueller, G.~Lakemeyer, and S.~S. Srinivasa, ``A generic robot database
  and its application in fault analysis and performance evaluation,'' in
  \emph{2012 IEEE/RSJ International Conference on Intelligent Robots and
  Systems}, Oct 2012, pp. 364--369.

\bibitem{Briegel2012}
H.~J. Briegel and G.~De~las Cuevas, ``Projective simulation for artificial
  intelligence,'' \emph{Scientific Reports}, vol.~2, pp. 400 EP --, May 2012,
  article.

\bibitem{14-toussaint-KOMO}
M.~Toussaint, ``{KOMO}: {N}ewton methods for k-order markov constrained motion
  problems,'' e-Print arXiv:1407.0414, 2014.

\bibitem{schaal2006dynamic}
S.~Schaal, ``Dynamic movement primitives-a framework for motor control in
  humans and humanoid robotics,'' in \emph{Adaptive motion of animals and
  machines}.\hskip 1em plus 0.5em minus 0.4em\relax Springer, 2006, pp.
  261--280.

\bibitem{ude2014orientation}
A.~Ude, B.~Nemec, T.~Petri{\'c}, and J.~Morimoto, ``Orientation in cartesian
  space dynamic movement primitives,'' in \emph{Robotics and Automation (ICRA),
  2014 IEEE International Conference on}.\hskip 1em plus 0.5em minus
  0.4em\relax IEEE, 2014, pp. 2997--3004.

\bibitem{libsvm}
C.-C. Chang and C.-J. Lin, ``{LIBSVM}: A library for support vector machines,''
  \emph{ACM Transactions on Intelligent Systems and Technology}, vol.~2, pp.
  27:1--27:27, 2011.

\bibitem{calli2015ycb}
B.~Calli, A.~Singh, A.~Walsman, S.~Srinivasa, P.~Abbeel, and A.~M. Dollar,
  ``The ycb object and model set: Towards common benchmarks for manipulation
  research,'' in \emph{Advanced Robotics (ICAR), 2015 International Conference
  on}.\hskip 1em plus 0.5em minus 0.4em\relax IEEE, 2015, pp. 510--517.

\end{thebibliography}


\end{document}